%% file: main.tex
\documentclass[10pt,twocolumn,letterpaper]{article}

\usepackage[pagenumbers]{CVPR/cvpr} % To force page numbers, e.g. for an arXiv version

\usepackage[dvipsnames]{xcolor}

\usepackage{hyperref}
\usepackage{url}
\usepackage{makecell}

\usepackage{times}
\usepackage{epsfig}
\usepackage{graphicx}
\usepackage{amsmath}
\usepackage{amssymb}
\usepackage{fontawesome5}
\usepackage{graphicx}
\usepackage{amsmath}
\usepackage{amssymb}
\usepackage{booktabs}
\usepackage[ruled,vlined]{algorithm2e}
\usepackage{algorithmic}
\usepackage{colortbl} % for coloring cells
\usepackage{booktabs} % for horizontal rules
\usepackage{multirow, booktabs, graphicx, xcolor, colortbl, arydshln, tikz}
\usetikzlibrary{tikzmark}
\usepackage{makecell}
\usepackage{siunitx} % for 'S' column type

\newcolumntype{g}{>{\columncolor{gray!30}}c}

\input{math_commands.tex}

\usepackage{tcolorbox}
\tcbuselibrary{most}

\usepackage{hyperref}

\definecolor{citecolor}{HTML}{0071bc}
\hypersetup{
    colorlinks=true,%
    citecolor=citecolor,%
    filecolor=citecolor,%
    linkcolor=citecolor,%
    urlcolor=citecolor
}

\tcbuselibrary{skins}

\definecolor{userbg}{RGB}{245, 245, 245}
\definecolor{userborder}{RGB}{210, 229, 255}
\definecolor{userfont}{RGB}{0, 0, 0}

\tcbset{
  enhanced,
  boxrule=1pt,
  colback=userbg,
  colframe=userborder,
  fonttitle=\bfseries,
  coltitle=userfont,
  boxsep=5pt,
  arc=5pt,
  outer arc=5pt,
  left=10pt,
  right=10pt,
  top=2pt,
  bottom=2pt,
  attach boxed title to top left={yshift=-0.25\baselineskip-1mm,xshift=2mm},
  boxed title style={
    colback=userborder,
    sharp corners,
    rounded corners=northeast,
    arc=3pt,
    outer arc=3pt,
    boxrule=0pt,
    boxsep=2pt,
  },
}
\usepackage{pgf}

\usepackage{multirow}
\usepackage{colortbl} % For color
\usepackage{booktabs}
\usepackage{tcolorbox} % for colored description boxes
\usepackage{enumitem}
\usepackage{graphicx}
\usepackage{multirow}
\usepackage{array}
\definecolor{listcolor}{RGB}{50,120,230}
\usepackage{arydshln}
\usepackage{url}
\newcommand{\thinparagraph}[1]{\vspace{0.1em}\par\noindent\textbf{#1}\enspace}
\newcommand{\ours}[1]{MMVP} % This definition replaces the content with "MVP"

\definecolor{cvprblue}{rgb}{0.21,0.49,0.74}
\def\paperID{6268} % *** Enter the Paper ID here
\def\confName{CVPR}
\def\confYear{2024}

\title{Eyes Wide Shut? Exploring the Visual Shortcomings of Multimodal LLMs}

\author{%
  Shengbang Tong\textsuperscript{1}\qquad Zhuang Liu\textsuperscript{2}
  \qquad 
  Yuexiang Zhai\textsuperscript{3}\qquad\\[1mm] Yi Ma\textsuperscript{3}\qquad Yann LeCun\textsuperscript{1}\qquad Saining Xie\textsuperscript{1}\\[3mm]
  \textsuperscript{1}New York University\qquad % 
  \textsuperscript{2}FAIR, Meta \qquad
  \textsuperscript{3}UC Berkeley
}

\begin{document}
\twocolumn[{
\maketitle
\vspace{-35pt}
\begin{center}
    \centering
    \captionsetup{type=figure}
    \includegraphics[width=1.0\textwidth]{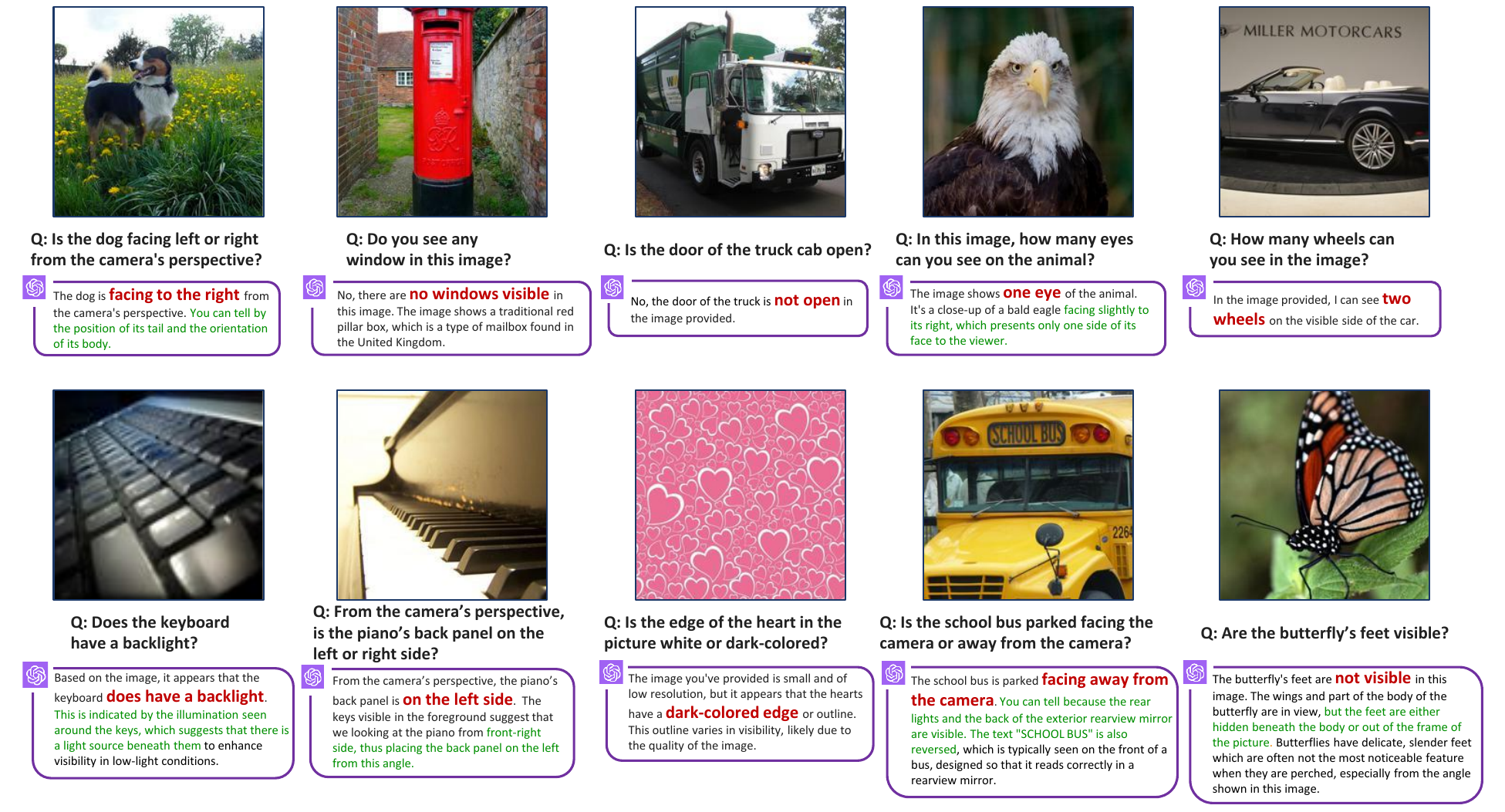}
  \vspace{-1.5em}
  \caption{Instances are systematically identified where the visual question answering (VQA) capabilities of GPT-4V~\citep{openai2023gpt4} fall short \text{\footnotesize{(\texttt{Date}\:\:\texttt{accessed}: \texttt{Nov}\:\texttt{04},\:\texttt{2023})}}. 
  Our research highlights scenarios in which advanced systems like GPT-4V struggle with seemingly simple questions due to inaccurate visual grounding. Text in \textbf{\textcolor{BrickRed}{red}} signifies an incorrect response, while text in \textbf{\textcolor{OliveGreen}{green}} represents hallucinated explanations for the incorrect answer. All the images referenced are sourced from ImageNet-1K and LAION-Aesthetic datasets.}
  \label{fig:teaser}
\end{center}
}]

\begin{abstract}
\input{sections/0_abstract}

\end{abstract}
\input{sections/1_intro}

\input{sections/2_benchmark}

\input{sections/3_visual_patterns_inclip}

\input{sections/4_experiments}
\vspace{-0.45cm}
\input{sections/5_more_related_works}
\input{sections/6_discussion}
\small{\noindent\textbf{Acknowledgements.} We thank Penghao Wu, Muzi Tao, Erik Jones, Michael Psenka, Daniel Yeh, Druv Pai, Chen Sun for helpful discussions and feedback. This work was supported in part through the NYU IT High Performance Computing resources, services, and staff expertise. This research is also supported by Intel, Google TRC program, the Google Cloud Research Credits program with the award GCP19980904, and an Amazon Research Award Fall 2023. The authors thank hyperbolic labs for supporting part of the experiments. All experiments and data processing were performed at NYU. 
{
    \small
    \bibliographystyle{CVPR/ieeenat_fullname}
    \bibliography{cvpr2024_conference}
}

\newpage
\input{sections/7_appendix}

\end{document}

%% file: math_commands.tex
\renewcommand{\mathbf}{\boldsymbol}

\makeatletter
\def\Ddots{\mathinner{\mkern1mu\raise\p@
\vbox{\kern7\p@\hbox{.}}\mkern2mu
\raise4\p@\hbox{.}\mkern2mu\raise7\p@\hbox{.}\mkern1mu}}
\makeatother
\numberwithin{equation}{section}

%% file: sections/0_abstract.tex
\vspace{-0.5cm}
Is vision good enough for language? 
Recent advancements in multimodal models primarily stem from the powerful reasoning abilities of large language models (LLMs). However, the visual component typically depends only on the instance-level contrastive language-image pre-training (CLIP). Our research reveals that the visual capabilities in recent MultiModal LLMs (MLLMs) still exhibit systematic shortcomings. To understand the roots of these errors, we explore the gap between the visual embedding space of CLIP and vision-only self-supervised learning. We identify ``CLIP-blind pairs'' – images that CLIP perceives as similar despite their clear visual differences. With these pairs, we construct the Multimodal Visual Patterns (\ours{}) benchmark. \ours{} exposes areas where state-of-the-art systems, including GPT-4V, struggle with straightforward questions across nine basic visual patterns, often providing incorrect answers and hallucinated explanations. We further evaluate various CLIP-based vision-and-language models and found a notable correlation between visual patterns that challenge CLIP models and those problematic for multimodal LLMs. As an initial effort to address these issues, we propose a Mixture of Features (MoF) approach, demonstrating that integrating vision self-supervised learning features with MLLMs can significantly enhance their visual grounding capabilities. Together, our research suggests visual representation learning remains an open challenge, and accurate visual grounding is crucial for future successful multimodal systems.

%% file: sections/1_intro.tex
\section{Introduction}

Multimodal Large Language Models (MLLMs)~\citep{gpt4v, Bard, liu2023visual, instructblip} have been rapidly developing in recent times. MLLMs integrate images into large language models (LLMs) and leverage the powerful abilities of LLMs~\citep{openai2023gpt4, touvron2023llama2, zheng2023judging}, showcasing remarkable proficiency in tasks such as image understanding, visual question answering, and instruction following. In particular, the recently released GPT-4V(ision)~\citep{gpt4v} has pushed performance to an unprecedented level~\citep{openai2023gpt4, yang2023dawn}.

Beneath the advancements of these models, we find there exists a notable weakness: they still exhibit visual shortcomings, some of which are surprisingly elementary and evident (see Figure~\ref{fig:teaser}). We ask: {\em Where do these problems originate? Is it a deficiency in visual modality, language understanding, or their alignment?} In this work, we suggest that these shortcomings observed in MLLMs might stem from a problem related to the \textbf{visual representations}.

At their core, most MLLMs~\citep{instructblip, liu2023visual, zhu2023minigpt} are built on \emph{pretrained} vision~\citep{radford2021learning, sun2023eva} and language~\citep{zhang2023llama, touvron2023llama2, zheng2023judging} models. These models are connected using various types of adapters~\citep{alayrac2022flamingo, li2023blip2, liu2023visual} to integrate the different modalities. A natural hypothesis is that any limitation in the pretrained vision models can cascade into the downstream MLLMs that adopt them. Studies have explored a similar issue for language. For example,~\citet{yuksekgonul2022and, tong2023mass} demonstrate that failure patterns in the pretrained text encoder~\citep{radford2021learning, raffel2020exploring} will lead to downstream failures in text-guided generative models~\citep{rombach2022high, jun2023shap}. 

On the vision side, most open-source MLLMs~\citep{alayrac2022flamingo, li2023blip2, liu2023visual} adopt the pretrained Contrastive Language-Image Pre-Training (CLIP) model~\citep{radford2021learning} as the visual encoder. We begin by identifying failure examples that CLIP struggles to encode properly (Section~\ref{sec: mllm_benchmark}).  Inspired by~\citet{tong2023mass}, we exploit the {\em erroneous agreements} in the embedding space. If two visually different images are encoded similarly by CLIP, then at least one of the images is likely ambiguously encoded. We call such a pair of images a {\em CLIP-blind} pair. To measure the visual similarity between images, we use a vision-only self-supervised encoder such as DINOv2~\citep{oquab2023dinov2}. In this context, \emph{CLIP-blind} pairs are images with similar CLIP embeddings but different DINOv2 embeddings. 

We discover that these CLIP-blind pairs indeed lead to errors in downstream MLLMs. With these pairs, We introduce the {\bf M}ulti{\bf M}odal {\bf V}isual {\bf P}atterns (\ours{}) benchmark. This benchmark is specifically designed to inquire about differences in CLIP-blind pairs and evaluate the visual abilities of \textit{state-of-the-art} MLLMs with straightforward questions. We evaluate a variety of open-source~\citep{liu2023improved, liu2023visual, instructblip, zhu2023minigpt} and closed-source models~\citep{openai2023gpt4, Bard} including GPT-4V~\citep{gpt4v}, and conduct a user study to measure human performance. The results show that MLLM models struggle with straight-forward visual questions.  Most of these models perform below the level of random guessing, with GPT-4V being the exception. Yet, even GPT-4V exhibits a considerable disparity in performance -- exceeding 50\% -- compared to human performance.

Having identified a large number of individual failure instances in MLLMs, we continue to study the systematic visual patterns in \ours{} which CLIP models struggle (Section~\ref{sec: analysis clip}). We summarize nine prevalent patterns of the CLIP-blind pairs in \ours{}, such as ``orientation'', ``counting'', and ``viewpoint'', which pose significant challenges for the CLIP vision encoder. Notice that there has been significant and ongoing progress in scaling up both training data and model size for CLIP~\citep{radford2021learning, sun2023eva, fang2023data, xu2023demystifying, zhai2023sigmoid}. We categorize examples from \ours{} into visual patterns to systematically assess whether scaling alone can mitigate these challenges. Our findings suggest that 7 out of the 9 identified visual patterns cannot be resolved by any large-scale CLIP-based models, indicating that model/data scaling alone is not sufficient. Moreover, we identify a strong correlation between the visual patterns that challenge CLIP models and the performance of MLLMs. If CLIP struggles with a particular visual pattern, such as ``orientation'', MLLMs will likely also fall short. This shows that the CLIP vision encoders could become a bottleneck in such systems.

Finally, we take a step towards improving the visual grounding of MLLMs. Since the visual shortcomings of MLLMs stem from their reliance on the CLIP model, we investigate the impact of integrating vision-centric representations into MLLMs (Section~\ref{sec: incorporate_dino}). Specifically, we explore ways to incorporate a vision-only self-supervised model, such as DINOv2~\citep{oquab2023dinov2}, to enhance the visual grounding capabilities of MLLMs. We refer to these techniques as Mixture-of-Features (MoF).  First, we linearly mix CLIP and DINOv2 features in different ratios, which we refer to as Additive-MoF (A-MoF). This process reveals that DINOv2 features are more effective in visual grounding, though they come at the cost of diminished instruction-following ability. To address this, we introduce Interleaved-MoF (I-MoF) that spatially mixes visual tokens from both CLIP and DINOv2 models. We find that this practice significantly enhances visual grounding while maintaining the instruction-following capabilities.

%% file: sections/2_benchmark.tex
\begin{figure*}[t]
  \centering
  \includegraphics[width=1.0\textwidth]{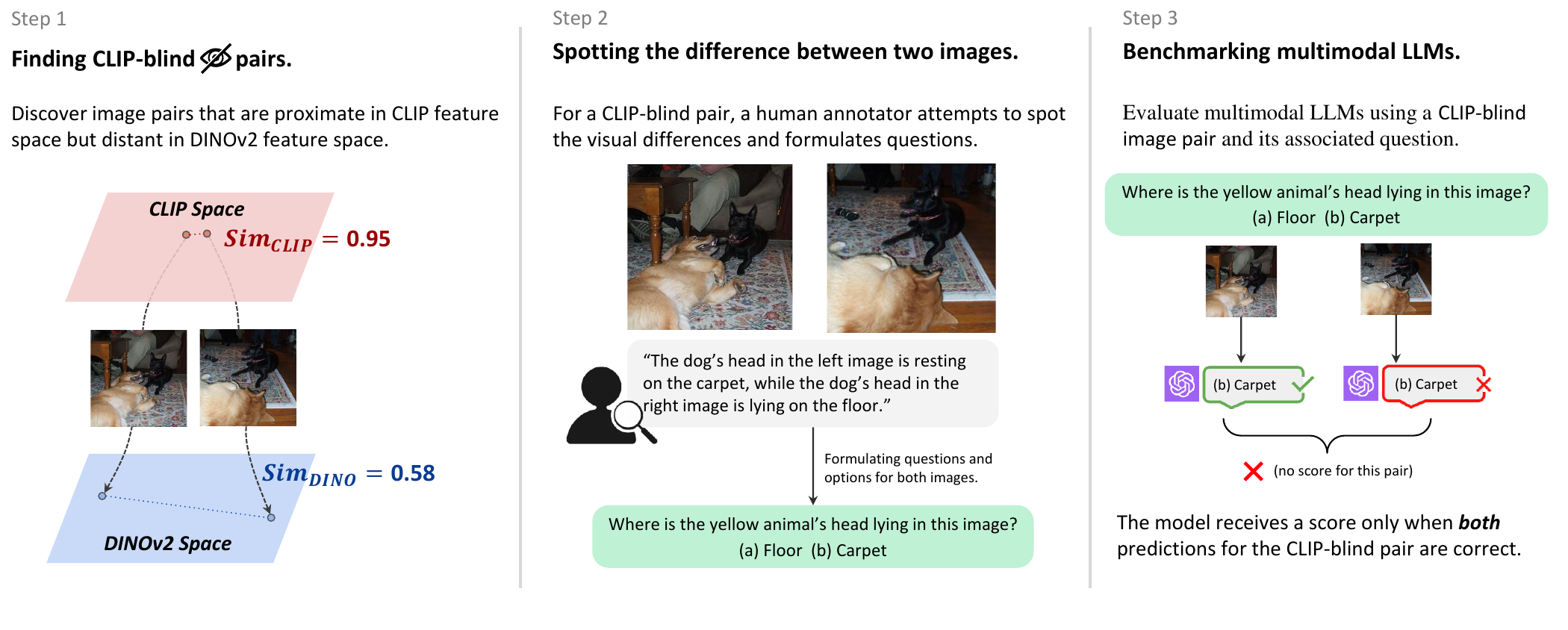}
  \caption{Constructing \ours{} benchmark via CLIP-blind pairs. \textbf{Left:} We start with finding CLIP-blind pairs that have similar CLIP embedding but different DINOv2 embedding. \textbf{Center:} We manually inspect the differences between pair-wise images and formulate questions based on the differences in the images. \textbf{Right:} We ask MLLMs the question alongside the CLIP-blind pair. The model receives a score only when both questions for the CLIP-blind pair are answered correctly. 
   }
  \label{fig:pipeline illustration}
\end{figure*}

\section{The Multimodal Visual Patterns (\ours{}) Benchmark} \label{sec: mllm_benchmark}
Currently, the majority of open-source MLLMs~\citep{liu2023visual, zhu2023minigpt, instructblip} use the {\em off-the-shelf} CLIP vision encoders to process images. In this section, we begin by identifying CLIP-blind pairs in the CLIP model (Section~\ref{sec: find adversarial in CLIP}). Subsequently, we construct the Multimodal Visual Patterns-MLLM (\ours{}-MLLM) benchmark using these CLIP-blind pairs (Section~\ref{sec: design benchmark}). We evaluate SOTA MLLMs including GPT-4V on the benchmark (Section~\ref{sec: evaluate benchmark}) and find that all the tested models struggle with simple questions on visual details. A visualization of this process is provided in Figure~\ref{fig:pipeline illustration}.

\subsection{Finding CLIP-blind Pairs} \label{sec: find adversarial in CLIP}
It is challenging to directly find instances (images) that the CLIP vision encoder struggles to encode ``properly''. To circumvent this issue, we extend the idea proposed in~\citet{tong2023mass} to automatically find blind pairs in vision models. The underlying principle is simple: if two images, despite having stark visual differences, are encoded similarly by the CLIP vision encoder, then one of them is likely encoded ambiguously (See Figure~\ref{fig:pipeline illustration} left for example). To measure the visual difference between two images, we examine the images' representations within a reference model: a vision-only self-supervised model trained without any language guidance, e.g., DINOv2~\citep{oquab2023dinov2}. These models are shown to capture more visual details and information~\citep{oquab2023dinov2, singh2023effectiveness}.

We take the corpus datasets, ImageNet~\citep{russakovsky2015imagenet} and LAION-Aesthetics~\citep{schuhmann2022laion}, to collect these CLIP-blind pairs. 

For each pair, we compute its CLIP embeddings using CLIP-ViT-L-14~\citep{dosovitskiy2021an, radford2021learning} model and their DINOv2 embeddings using DINOv2-ViT-L-14~\citep{dosovitskiy2021an, oquab2023dinov2} model. We return pairs such that the cosine similarity exceeds 0.95 for CLIP embeddings and less than 0.6 for DINOv2 embeddings. 

\subsection{Designing Benchmark from CLIP-blind Pairs}
\label{sec: design benchmark}
We introduce the Multimodal Visual Patterns (\ours{}) benchmark, and a Visual Question Answering (VQA) benchmark. Utilizing the collected CLIP-blind pairs, we carefully design 150 pairs with 300 questions. For each CLIP-blind pair of images, we manually pinpoint the visual details that the CLIP vision encoder overlooks (see the middle of Figure~\ref{fig:pipeline illustration}) and craft questions that probe these visual details, for example ``Is the dog facing left or right?'' (See the right of Figure~\ref{fig:pipeline illustration} and more examples in Figure~\ref{fig:example_questions}). The primary goal is to determine whether MLLM models would fail when posed with these seemingly basic questions and overlook critical visual details. Hence, the questions are intentionally straightforward and unambiguous. 

\definecolor{rose}{HTML}{FF9999}

\begin{figure*}[h]
  \centering
  \includegraphics[width=0.98\textwidth]{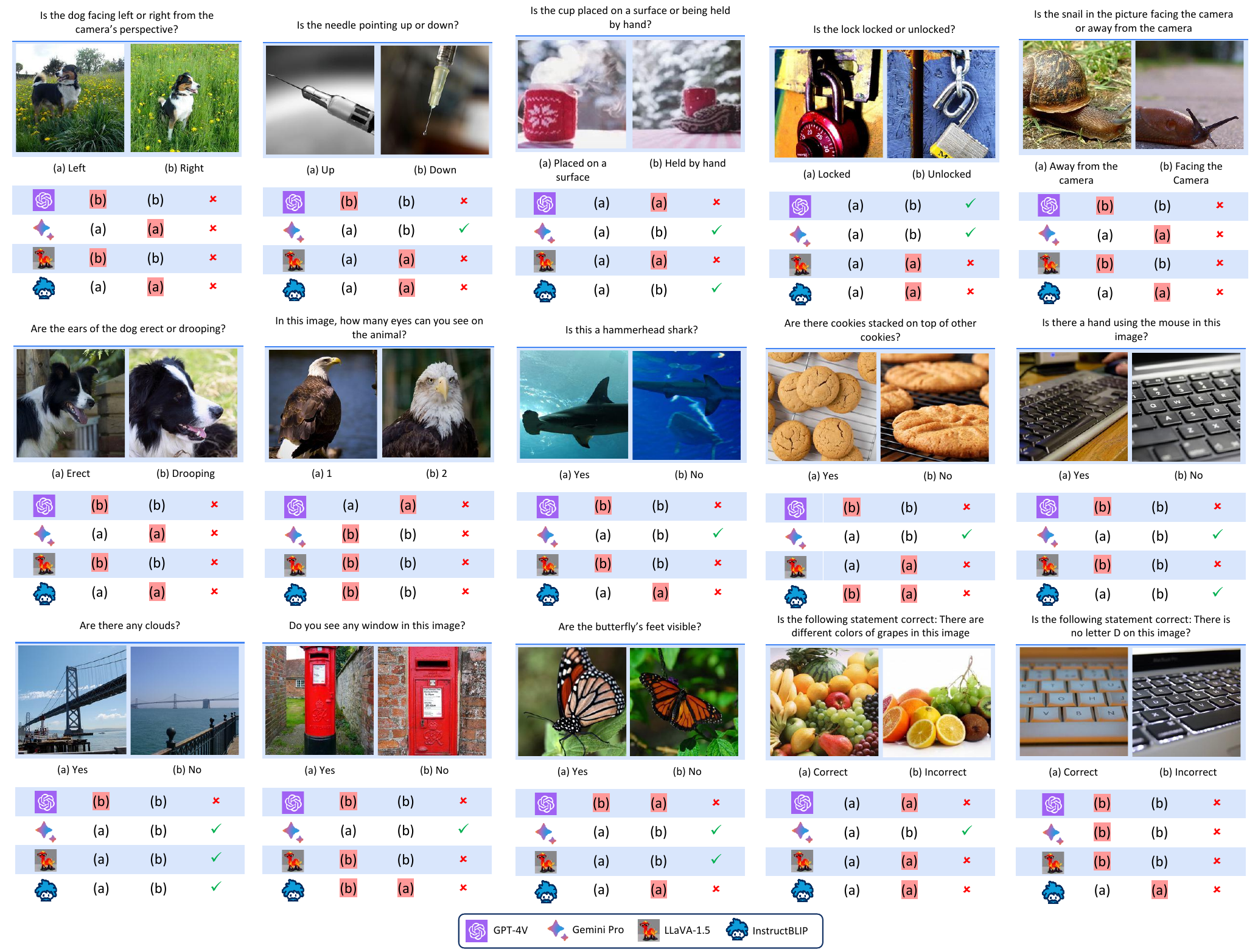}
  \caption{\textbf{Examples of Questions in the \ours{} benchmark.} Incorrect answers are shaded in \colorbox{rose}{red}.
  A model is considered correct only if it answers both questions in a pair correctly. Both leading closed-source models (GPT-4V, Gemini) and open-source models (LLaVA-1.5, InstructBLIP) fail these simple visual questions. (See Appendix~\ref{appendix: full benchmark} for all the questions in \ours{} benchmark.)}
  \label{fig:example_questions}
  \vspace{-0.5cm}
\end{figure*}

\subsection{Benchmark Results} 
\label{sec: evaluate benchmark}
We assess the questions on \textit{SOTA} open-source models (LLaVA-1.5~\citep{liu2023visual}, InstructBLIP~\citep{instructblip}, Mini-GPT4~\citep{zhu2023minigpt}) and closed-source models (GPT-4V~\citep{gpt4v}, Gemini~\citep{Gemini}, Bard~\citep{Bard}) 
We leave details of how we access the model in Appendix~\ref{appendix: access the model}. In our evaluation, each question is queried independently, eliminating any biases from chat histories. We also evaluate human performance through a user study where users are presented with 300 questions in a randomized sequence. For any given pair of images, we consider a pair of images to be correctly answered if both the questions associated with the pair are answered accurately.

\begin{figure}[t]
  \centering
    \hspace{-0.5cm}  % Adjust the value as needed
  \includegraphics[width=0.47\textwidth]{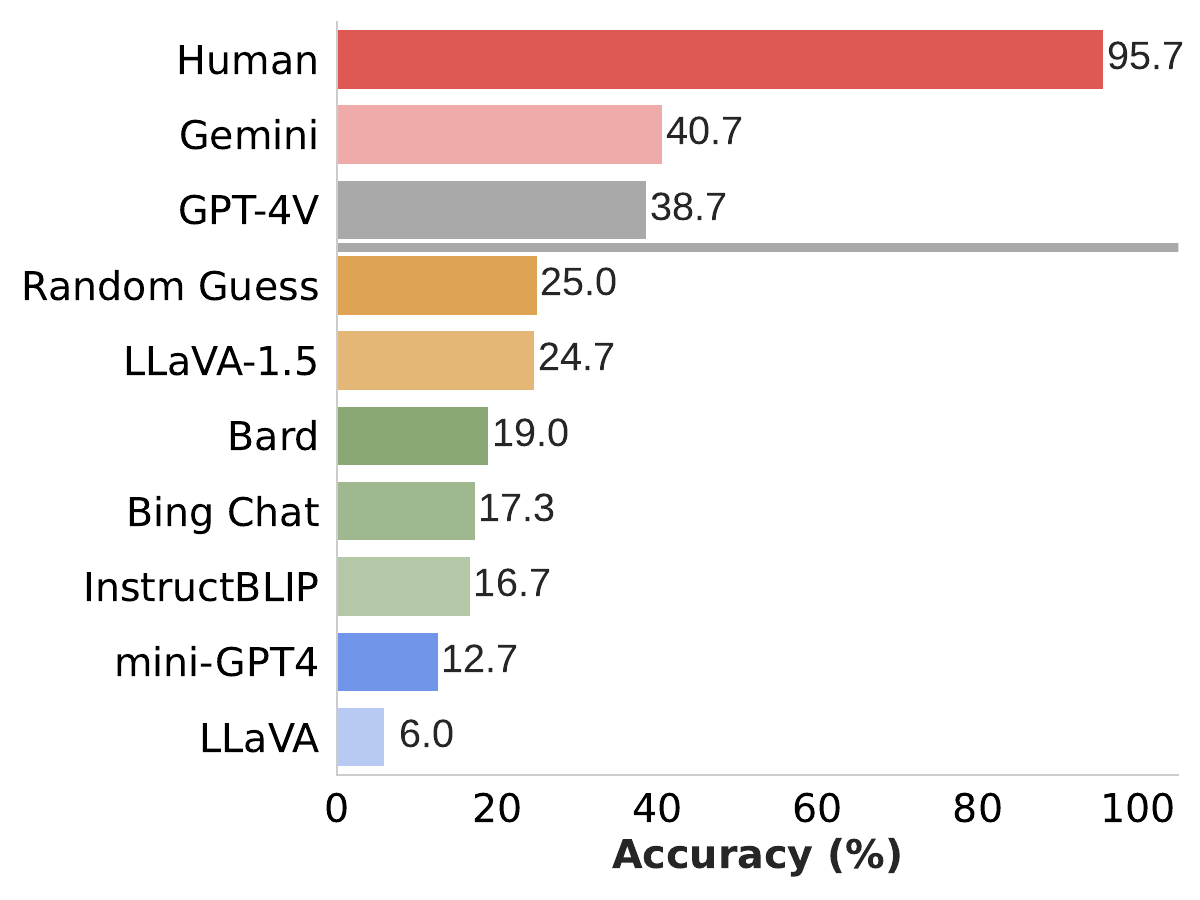}
  \caption{\textbf{Benchmark results of current \textit{SOTA} MLLM models and humans.} We evaluate benchmark questions for current \textit{SOTA} MLLM models and human performances through user studies. }
  \label{fig:benchmark_results}
  \vspace{-1cm}
\end{figure}
\definecolor{darkgreen}{rgb}{0.0, 0.5, 0.0} % Adjust RGB values as needed
\definecolor{darkred}{RGB}{175, 29, 33}

\begin{figure*}[t]
  \centering
  \includegraphics[width=1.0\textwidth]{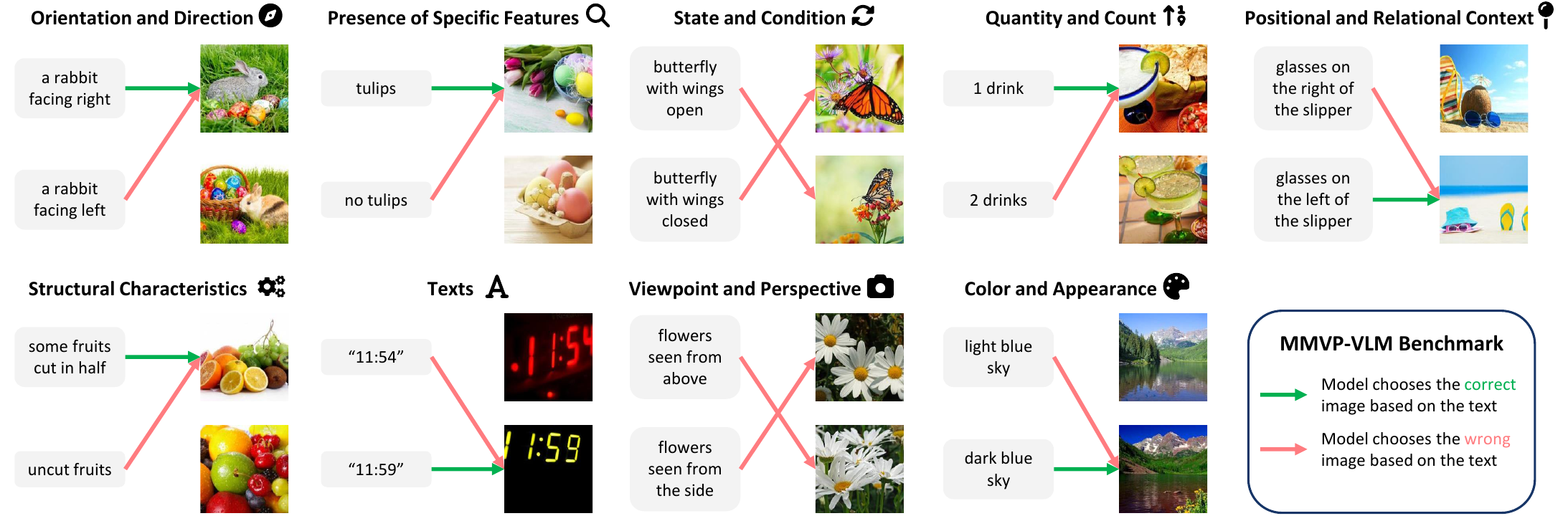}
  \caption{\textbf{Examples from \ours{}-VLM}. 
  \ours{}-VLM consists of image pairs across nine visual patterns. 
  The examples in the figure are from EVA01 ViT-g-14 model \citep{sun2023eva}, one of the largest CLIP models that also fails to choose the right image given the text description.} 
  \label{fig:winoground-v demo}
  \vspace{-0.5cm}
\end{figure*}

\paragraph{Human study confirms questions are straightforward.}
As shown in Figure~\ref{fig:benchmark_results}, human participants accurately answer an average of 95.7\% of the questions. 
This high accuracy rate underscores the ease of the questions. More details can be found in Appendix~\ref{appendix: human study}.

\paragraph{Current MLLMs struggle with visual details.} As shown in Figure~\ref{fig:benchmark_results}, there is a significant performance gap between human and MLLM models, despite the latter often demonstrating impressive results~\citep{bubeck2023sparks, li2023evaluating}. Models except GPT-4V and Gemini, scored below random guess level (25\%). Most advanced GPT-4V and Gemini also face challenges in addressing basic visual grounding questions. Figures~\ref{fig:teaser} and~\ref{fig:example_questions} provide examples of errors made by models. The outcomes suggest that irrespective of model size or training data, struggle with visual details. 

We have also conducted an ablation study, such as swapping options and changing notations in the question formulation (see Appendix~\ref{appendix: ablation study on questions} for more details), to further confirm that this poor performance stems from visual incapability, not hallucination in the language models.

%% file: sections/3_visual_patterns_inclip.tex
\section{Systematic Failures in CLIP} \label{sec: analysis clip}

In the previous section, we identify CLIP-blind pairs and use them to find failures in MLLMs. Here, we delve deeper into these pairs to investigate  (i) systematic visual patterns emerged from CLIP-blind pairs (Section~\ref{sec: visual patterns}), (ii) whether these visual patterns pose challenges for CLIP-based models with massive scaling up (Section~\ref{sec: vlm benchmark}), and (iii) the correlation between failure patterns in CLIP models and those in MLLMs (Section~\ref{sec: correlation}).

\subsection{Visual Patterns in CLIP-blind Pairs} \label{sec: visual patterns}
Having identified the CLIP-blind pairs, we summarize systematic visual patterns that the CLIP vision encoders might consistently misinterpret. It is too abstract to directly capture systematic visual patterns in the CLIP-blind pairs. Therefore, we turn to the questions and options from the \ours{} benchmark. With these questions, we transform abstract visual patterns in images into clearer, language-based descriptors that are easier to categorize. 

In this work, we use  GPT-4~\citep{openai2023gpt4} to categorize general patterns by prompting it with the following: 
\begin{tcolorbox}[title=User]
    \small
    I am analyzing an image embedding model. Can you go through the questions and options, trying to figure out some general patterns that the embedding model struggles with? Please focus on the visual features and generalize patterns that are important to vision models
    
    [\ours{} Questions and Options]
\end{tcolorbox}
We identify 9 visual patterns:

\begin{tabular}{ll}
    %\toprule
    \textbf{\faCompass} & Orientation and Direction \\
    \textbf{\faSearch} & Presence of Specific Features \\
    \textbf{\faSync} & State and Condition \\
    \textbf{\faSortNumericUp} & Quantity and Count \\
    \textbf{\faMapPin} & Positional and Relational Context \\
    \textbf{\faPalette} & Color and Appearance \\
    \textbf{\faCogs} & Structural and Physical Characteristics \\
    \textbf{\faFont} & Text \\
    \textbf{\faCamera} & Viewpoint and Perspective \\
    %\bottomrule
\end{tabular}

These visual patterns suggest that CLIP vision encoders overly focus on high-level semantic understanding, overlooking intricate details of the visual world. 
Full descriptions of the visual patterns can be found in Appendix~\ref{appendix: visual patterns}. 
% Define shades of green
\definecolor{LightGreen}{HTML}{ccffcc}
\definecolor{Green}{HTML}{99ff99}
\definecolor{DarkGreen}{HTML}{66cc66}

% Define shades of red
\definecolor{LightRed}{HTML}{ffcccc}
\definecolor{Red}{HTML}{ff9999}
\definecolor{DarkRed}{HTML}{ff6666}

% Define shades of blue
\definecolor{LightBlue}{HTML}{cce0ff}
\definecolor{Blue}{HTML}{99ccff}
\definecolor{DarkBlue}{HTML}{668cff}
\definecolor{LightTeal}{HTML}{B3FFFF}
\definecolor{MediumTeal}{HTML}{66FFFF}
\definecolor{DarkTeal}{HTML}{33CCCC}
\definecolor{LightYellow}{HTML}{FFFFCC}
\definecolor{MediumYellow}{HTML}{FFFF99}
\definecolor{DarkYellow}{HTML}{FFFF66}
\definecolor{lightgray}{gray}{0.9}

\begin{table*}[t]
    \centering
    \small % Reducing font size to fit table
    \setlength\tabcolsep{5pt} % Default value: 6pt
    \begin{tabular}{l:ccc:ccccccccc:c}
        & \multirow{2}{*}{\shortstack{Image\\Size}} & \multirow{2}{*}{\shortstack{Params \\ (M)}} & \multirow{2}{*}{\shortstack{IN-1k \\ ZeroShot}} & \multirow{2}{*}{\faCompass}  &  \multirow{2}{*}{\faSearch} & \multirow{2}{*}{\faSync} & \multirow{2}{*}{\faSortNumericUp} & \multirow{2}{*}{\faMapPin} & \multirow{2}{*}{\faPalette}  & \multirow{2}{*}{\faCogs}  & \multirow{2}{*}{\faFont} & \multirow{2}{*}{\faCamera}  & \multirow{2}{*}{\shortstack{\ours{} \\ Average}}
         \\
         \\
        \Xhline{1pt} 
        
        OpenAI ViT-L-14 \citep{radford2021learning} & \cellcolor{LightBlue}224$^2$ & 427.6 & 75.5 & 13.3 & 13.3 & 20.0 & 20.0 & 13.3 & 53.3 & 20.0 & 6.7 & 13.3 & 19.3 \\
        OpenAI ViT-L-14 \citep{radford2021learning}& \cellcolor{Blue}336$^2$ & 427.9 & 76.6 & 0.0 & 20.0 & 40.0 & 20.0 & 6.7 & 20.0 & 33.3 & 6.7 & 33.3 &  20.0 \\
        SigLIP ViT-SO-14 \citep{zhai2023sigmoid}& \cellcolor{LightBlue}224$^2$ & 877.4 & 82.0 & \cellcolor{lightgray}26.7 & 20.0 & 53.3 & \cellcolor{lightgray}40.0 & 20.0 & \cellcolor{lightgray}66.7 & 40.0 & 20.0 & \cellcolor{lightgray}53.3 & 37.8 \\
        SigLIP ViT-SO-14 \citep{zhai2023sigmoid}& \cellcolor{Blue}384$^2$ & 878.0 & 83.1 &20.0 & \cellcolor{lightgray}26.7 & 60.0 & 33.3 & 13.3 & \cellcolor{lightgray}66.7 & 33.3 & \cellcolor{lightgray}26.7 & \cellcolor{lightgray}53.3 & 37.0 \\
        DFN ViT-H-14~\citep{fang2023data}& \cellcolor{LightBlue}224$^2$ & 986.1 & 83.4 & 20.0 & \cellcolor{lightgray}26.7 & \cellcolor{lightgray}73.3 & 26.7 & 26.7 & \cellcolor{lightgray}66.7 & \cellcolor{lightgray}46.7 & 13.3 & \cellcolor{lightgray}53.3 & \cellcolor{lightgray}39.3 \\

        DFN ViT-H-14~\citep{fang2023data}& \cellcolor{Blue}378$^2$ & 986.7 & \cellcolor{lightgray}84.4 & 13.3 & 20.0 & 53.3 & 33.3 & 26.7 & \cellcolor{lightgray}66.7 & 40.0 & 20.0 & 40.0 & 34.8 \\
        
        MetaCLIP ViT-L-14 \citep{xu2023demystifying}& 224$^2$ & \cellcolor{LightGreen}427.6 & 79.2 & 13.3 & 6.7 & 66.7 & 6.7 & \cellcolor{lightgray}33.3 & 46.7 & 20.0 & 6.7 & 13.3 & 23.7 \\
        MetaCLIP ViT-H-14 \citep{xu2023demystifying}& 224$^2$ & \cellcolor{Green}986.1 & 80.6 & 6.7 & 13.3 & 60.0 & 13.3 & 6.7 & 53.3 & 26.7 & 13.3 & 33.3 & 25.2 \\
        EVA01 ViT-g-14 \citep{sun2023eva}& 224$^2$ & \cellcolor{Green}1136.4 & 78.5 & 6.7 &\cellcolor{lightgray} 26.7 & 40.0 & 6.7 & 13.3 & \cellcolor{lightgray}66.7 & 13.3 & 13.3 & 20.0 & 23.0 \\
        EVA02 ViT-bigE-14+ \citep{sun2023eva}& 224$^2$ & \cellcolor{DarkGreen} 5044.9 & 82.0  & 13.3 & 20.0 & 66.7 & 26.7 & 26.7 & \cellcolor{lightgray}66.7 & 26.7 & 20.0 & 33.3 & 33.3  \\
    \end{tabular}
    \caption{Performance of various CLIP based models on different visual patterns in \ours{}-VLM benchmark. Models \colorbox{LightBlue}{scaled up in resolution} show minimal improvement, whereas a slight advantage is observed when \colorbox{LightGreen}{scaling up the network.}  For each visual pattern, ImageNet-1k Zero-shot accuracy and \ours{} average, we use \colorbox{lightgray}{light gray} to highlight the best performance. For most of the visual patterns, all CLIP-based methods show struggle, as evident from the scores. We use symbols for visual patterns due to space limit:
    \textbf{\faCompass}: Orientation and Direction, \textbf{\faSearch}: Presence of Specific Features, \textbf{\faSync}: State and Condition, \textbf{\faSortNumericUp}: Quantity and Count, \textbf{\faMapPin}: Positional and Relational Context, \textbf{\faPalette}: Color and Appearance, \textbf{\faCogs}: Structural and Physical Characteristics, \textbf{\faFont}: Texts, \textbf{\faCamera}: Viewpoint and Perspective.
    }
    \label{tab:winoground-v results}
\end{table*}

\subsection{The \ours{}-VLM Benchmark} \label{sec: vlm benchmark}

CLIP-based models have developed rapidly since the introduction in the first paper~\citep{radford2021learning}. We want to test whether these visual patterns still impose challenges to the more recent CLIP models~\citep{fang2023data, sun2023eva, zhai2023sigmoid, xu2023demystifying}, which significantly scale up in terms of training data and model size. In doing so, we introduce a new benchmark: \ours{}-VLM to systematically study if CLIP models handle this visual pattern well. 

We distill a subset of questions from the \ours{} benchmark into simpler language descriptions and categorize them into visual patterns. To maintain a balanced number of questions for each visual pattern, we add a few questions, if needed, to ensure that each visual pattern is represented by 15 text-image pairs.
Examples of pairs are shown in Figure~\ref{fig:winoground-v demo}.  A pair is deemed correctly answered if the model can accurately match both image-text combinations. 

We evaluate \ours{}-VLM on a variety of CLIP models~\citep{radford2021learning, sun2023eva, fang2023data, xu2023demystifying, zhai2023sigmoid}. These models vary in aspects like size, training data, and methodology. As evidenced in Table~\ref{tab:winoground-v results}, increasing network size and training data only aids in identifying two visual patterns -- ``color and appearance'' and ``state and condition''. The rest of the visual patterns continue to challenge all CLIP-based models. We also find that the ImageNet-1k zero-shot accuracy is not a definitive indicator of a model's performance regarding visual patterns. This underscores the necessity for additional evaluation metrics, such as \ours{}-VLM, to accurately assess the model's capabilities in areas beyond image classification.

\subsection{How CLIP's Errors Affect MLLMs} \label{sec: correlation}
\begin{figure}[t]
    \centering
  \includegraphics[width=0.49\textwidth]{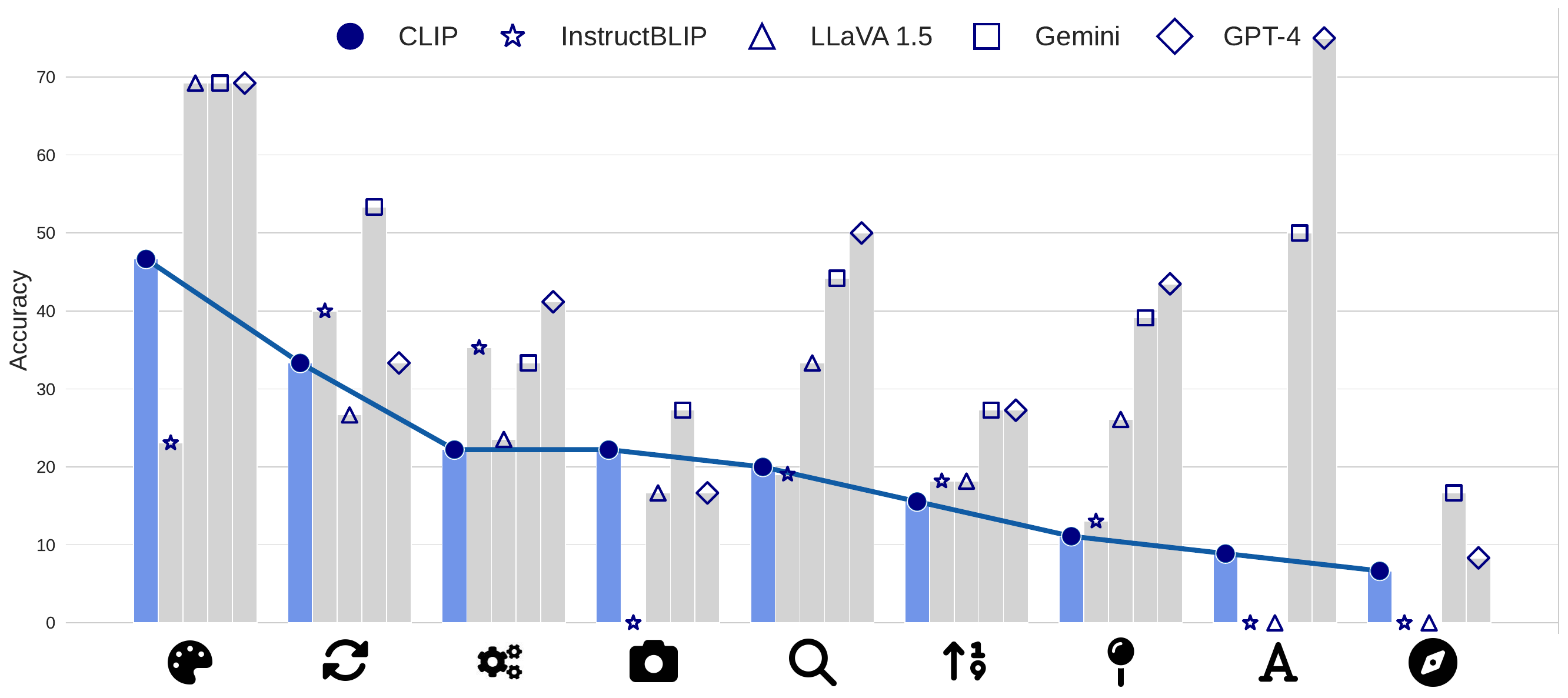}
  \caption{\textbf{CLIP and MLLM's performance on visual patterns.} If CLIP performs poorly on a visual pattern such as `` \textbf{\faCompass} orientation'', MLLMs also underperform on the visual pattern. 
   }
  \label{fig:correlation between MVP and WinoGround}
  \vspace{-0.5cm}
\end{figure}

After analyzing the visual patterns that CLIP models struggle with, we pose the following question: Is there a correlation between the underperformance of CLIP and MLLMs' visual incapability? To explore this, we categorize questions from \ours{} into these visual patterns summarized and calculate each MLLM's performance on these patterns. 

In Figure~\ref{fig:correlation between MVP and WinoGround}, we plot CLIP’s performance and MLLMs' performance for each visual pattern. When the CLIP vision encoder underperforms on a certain visual pattern, the MLLM tends to exhibit similar shortcomings. Open-source models such as LLaVA 1.5~\citep{liu2023improved} and InstructBLIP~\citep{instructblip} that explicitly use the CLIP vision encoder display a strong correlation in performance.

Further,  we calculate the Pearson Correlation Coefficient between the CLIP model and MLLM's performance on each visual pattern. Results show that LLaVA 1.5 and InstructBLIP all possess a coefficient score greater than 0.7. This high score indicates a strong correlation that weaknesses in visual pattern recognition in the CLIP model are transferred to MLLMs. More details on the Pearson Correlation Coefficient can be found in Appendix~\ref{appendix: correlation}.

%% file: sections/4_experiments.tex
\section{Mixture-of-Features (MoF) for MLLM} \label{sec: incorporate_dino}

\begin{figure*}[t]
    \centering
  \includegraphics[width=0.99\textwidth]{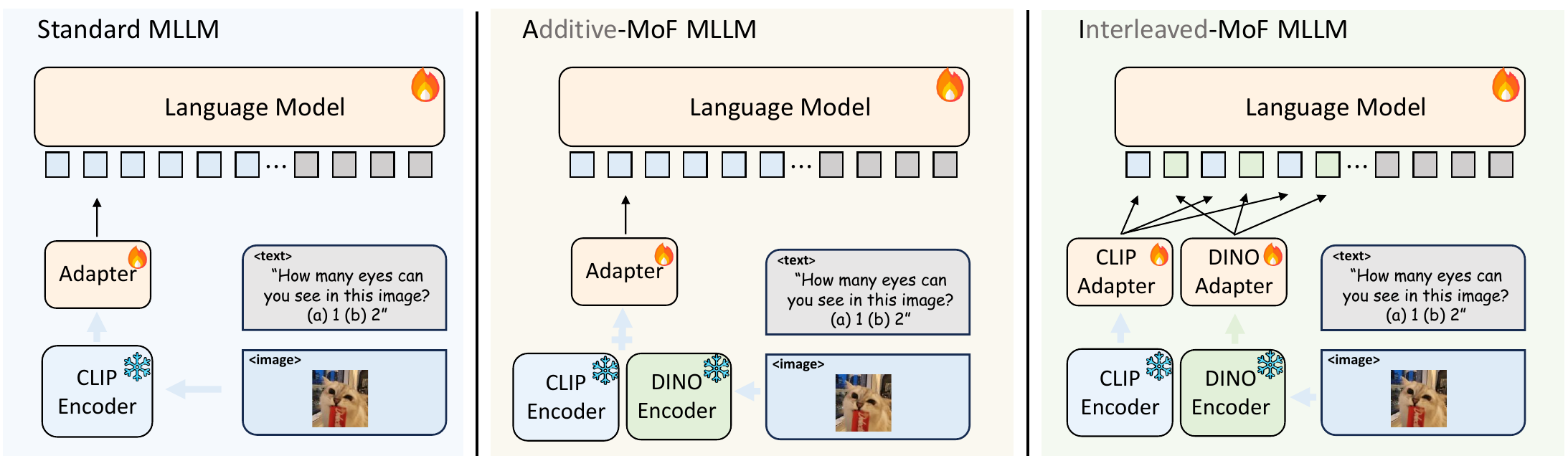}
  \caption{\textbf{Different Mixture-of-Feature (MoF) Strategies in MLLM.} \textit{Left}: Standard MLLM that uses CLIP as \textit{off-the-shelf} pretrained vision encoder; \textit{Middle}: Additive-MoF (A-MoF) MLLM: Linearly mixing CLIP and DINOv2 features before the adapter; \textit{Right}: Interleaved-MoF (I-MoF MLLM) Spatially interleaving CLIP visual tokens and DINOv2 visual tokens after the adapter. }
  \label{fig: MOF MLLM}
  \vspace{-0.5cm}
\end{figure*}

Based on our exploration in earlier sections, a natural question arises: \textit{If open-sourced MLLM's visual shortcomings come from the CLIP vision encoder, how do we build a more competent visual encoder?} In this section, we take initial steps to answer the question by studying Mixture-of-Features (MoF). We start with additive MoF that mixes CLIP features and vision-only SSL model features. Results show that each encoder presents unique advantages and limitations when employed as the pretrained model in MLLM (Section~\ref{sec: tradeoff between CLIP and DINO}). We subsequently propose Interleaved MoF that integrates the features from both CLIP and SSL into MLLM to enhance visual grounding without compromising the model's ability to follow instructions (Section~\ref{sec: spatial interleave}). 

\subsection{Experiment Setting} \label{sec: Experiment Setting}
We adopt LLaVA~\citep{liu2023visual, liu2023improved} as the framework to study visual encoders in MLLM. LLaVA uses a pretrained CLIP encoder and trains an adapter to align visual tokens with language tokens in the LLM. (See left side of Figure~\ref{fig: MOF MLLM}).  We use DINOv2~\citep{oquab2023dinov2} as the vision-only SSL model in our work because it is currently the most scalable vision-only model. Our exploration includes the use of two visual encoders: CLIP-ViT-L-14 \citep{radford2021learning} and DINOV2-ViT-L-14~\citep{oquab2023dinov2}. To ensure consistent and fair comparisons, we train and finetune our model with the same experiment setting in LLaVA. We include the additional experimental details in Appendix~\ref{appendix: experiment details}.

\subsection{Additive MoF}
\label{sec: tradeoff between CLIP and DINO}
We add a pretrained DINOv2 encoder into MLLM and mix the CLIP pretrained encoder with it. We use a coefficient $\alpha$ to control the portion of CLIP features and $1-\alpha$ to control the amount of DINOv2 features and {\em linearly} add them together (See middle part of Figure~\ref{fig: MOF MLLM} for visualization).

We evaluate the model's visual grounding ability by the \ours{} proposed earlier in Section~\ref{sec: mllm_benchmark} and the model's instruction-following capability by LLaVA benchmark introduced in~\citet{liu2023visual}. Initially, we conduct five experiments where we linearly transition from using 100\% CLIP features to 100\% DINOv2 features. In these tests, the DINOv2 feature proportions are set at $\{0.00, 0.25, 0.50, 0.75, 1.00\}$. To further verify the observed trends, we introduce two additional experiments with DINOv2 proportions of $\{0.625, 0.875\}$. Our findings, presented in Table~\ref{tab:interpolate Results}, reveal two insights:
\begin{enumerate}
    \item As the proportion of DINOv2 features increases, MLLM exhibits a decline in its instruction-following capability. Notably, there is a sharp decrease when the DINOv2 proportion reaches 87.5\%.
    \item A higher proportion of DINOv2 features enhances the model's visual grounding capability, but this advantage diminishes when the DINOv2 proportion surpasses 0.75, at which point instruction-following is notably impaired.
\end{enumerate}
Hence, if we were to add DINOv2 features or completely replace CLIP with DINOv2, it would result in a trade-off between visual grounding and instruction-following. A higher proportion of DINOv2 features improves the model's visual perception at the expense of its ability to follow linguistic instructions, while CLIP features enhance language comprehension but reduce visual grounding.

\newcommand{\plusvalue}[1]{\hspace{0.3em}\textcolor{darkgreen}{(+#1)}}
\newcommand{\minusvalue}[1]{\hspace{0.3em}\textcolor{red}{(-#1)}}

\definecolor{darkgreen}{rgb}{0.0, 0.5, 0.0}

\definecolor{lightgray}{gray}{0.9}

\begin{table}[h!]
    \centering
    \small
    \setlength\tabcolsep{6pt} % Default value: 6pt

    \begin{tabular}{l l l l}  
        method & SSL ratio & \ours{} & LLaVA \\
        \toprule
        LLaVA & 0.0 & 5.5 & \cellcolor{lightgray}\textbf{81.8} \\
        \midrule
         %\cdashline{1-4}[4pt/2pt]
        %\cline{1-1}
        \multirow{6}{*}{\shortstack{LLaVA  \\ + A-MoF}} & 0.25 & 7.9\text{ }   {\tiny\textbf{\plusvalue{2.4}}} & 79.4{\tiny\textbf{\minusvalue{2.4}}} \\
        & 0.5 & 12.0{\tiny\textbf{\plusvalue{6.5}}} & 78.6{\tiny\textbf{\minusvalue{3.2}}} \\
        & 0.625 & 15.0{\tiny\textbf{\plusvalue{9.5}}} & 76.4{\tiny\textbf{\minusvalue{5.4}}} \\
        & \cellcolor{lightgray}0.75 & \cellcolor{lightgray}\textbf{18.7}{\tiny\textbf{\plusvalue{13.2}}} & 75.8{\tiny\textbf{\minusvalue{6.0}}}\\
        & 0.875 & 16.5{\tiny\textbf{\plusvalue{11.0}}} & 69.3{\tiny\textbf{\minusvalue{12.5}}} \\
        & 1.0 & 13.4{\tiny\textbf{\plusvalue{7.9}}} & 68.5{\tiny\textbf{\minusvalue{13.3}}} \\
    \end{tabular}
    \caption{\textbf{Empirical Results of Additive MoF.} We use DINOv2 as the image SSL model in our work. With more DINOv2 features added, there is an improvement in visual grounding, while a decline in instruction following ability. }
    \label{tab:interpolate Results}
\end{table}

\subsection{Interleaved MoF} \label{sec: spatial interleave}
We propose interleaved MoF to leverage advantages from both CLIP and DINOv2 embeddings to enhance image representation. An image concurrently passes into CLIP and DINOv2 encoders, and the resulting embeddings are individually processed by adapters. We take the processed features from CLIP and DINOv2 and interleave them while maintaining their original spatial order. We then feed the interleaved features to LLM (See right part of Figure~\ref{fig: MOF MLLM}).
\definecolor{MidGreen}{HTML}{AAFFAA}

\begin{table}[h!]
    \centering
    \footnotesize 
    \setlength\tabcolsep{5pt} % Default value: 6pt
    \begin{tabular}{l c c l l l }  
        %\toprule
         method & res & \#tokens & \ours{} & LLaVA & POPE
        \\
        \toprule
        LLaVA  & 224$^2$ & 256 & 5.5 & 81.8 & 50.0 \\
        LLaVA & 336$^2$ & \cellcolor{Green}576 & \cellcolor{lightgray}6.0 & 81.4 & 50.1\\
        LLaVA + I-MoF  & 224$^2$ & \cellcolor{MidGreen}512 & \cellcolor{lightgray}16.7{\textbf{\tiny{\plusvalue{10.7}}}} & 82.8 & 51.0
        \\
        \midrule
        LLaVA$^{1.5}$  & 336$^2$ & \cellcolor{Green}576 & \cellcolor{lightgray}24.7 & 84.7 & 85.9\\

        LLaVA$^{1.5}$ + I-MoF & 224$^2$ & \cellcolor{MidGreen}512 & \cellcolor{lightgray}28.0{\textbf{\tiny{\plusvalue{3.3}}}} & 82.7 & 86.3
    \end{tabular}
    \caption{\textbf{Empirical Results of Interleaved MoF.} Interleaved MoF improves visual grounding while maintaining same level of instruction following ability. }
    \label{tab:Spatial Results}
\end{table}

We summarize the results in Table~\ref{tab:Spatial Results}. Under the LLaVA setting, interleave MoF significantly enhances visual grounding, with a 10.7\% increase observed in \ours{}, without compromising the model's ability to follow instructions. This experiment is replicated with the LLaVA-1.5 setting and under various image resolution settings, yielding similar enhancements in performance. We also evaluate on POPE~\citep{li2023evaluating} which is designed to test hallucination in visual grounding. Interleaved-MoF also shows consistent improvement against the original LLaVA models. Merely increasing the image resolution, and consequently, the number of tokens does not boost visual grounding capabilities. Instead, it is the interleaving of MoF between vision-only SSL models and VLM models that leads to improved performance in visual grounding tasks. We conduct more experiments using MAE or MoCoV3 as vision-only SSL models in I-MoF and show similar improvements in visual grounding tasks in Appenfix~\ref{appendix: mae&moco}.
We also evaluated Interleaved MoF on additional benchmarks such as MM-Bench~\citep{liu2023mmbench} and GQA~\citep{hudson2019gqa}, finding that Interleaved MoF achieves similar performance on these benchmarks. Please refer to Appendix \ref{appendix: more benchmarks} for more results on these benchmarks.

%% file: sections/5_more_related_works.tex
\section{Related Works}
\thinparagraph{Multimodal LLMs.}
We study the limitations of Multimodal LLMs~\citep{gpt4v, Bard, liu2023improved, liu2023visual, instructblip} and explore possible ways to improve these models. Multimodal LLMs build from pretrained Large Language Models~\citep{openai2023gpt4, anil2023palm, touvron2023llama, touvron2023llama2, zheng2023judging} and CLIP vision encoder~\citep{radford2021learning,sun2023eva}. These systems then use an adapter, such as MLPs~\citep{liu2023improved, liu2023visual}, Q-Former~\citep{li2023blip2, instructblip}, and gated attention~\citep{alayrac2022flamingo, laurenccon2023obelisc}, to integrate the pretrained CLIP vision encoder into LLMs. More recently, instructBLIP~\citep{instructblip}, LLaVA-1.5~\citep{liu2023improved} highlight the importance of high-quality training data. Yet, there is a scarcity of research focusing on the impact of visual encoders, which is an important gap our work aims to address through a systematic study. 

\thinparagraph{Evaluating Multimodal LLMs.}
\ours{} assesses MLLMs using a set of simple yet critical Visual Question Answering (VQA) questions constructed from CLIP-blind pairs.
Previous benchmarks such as TextVQA~\citep{singh2019towards}, VQAv2~\citep{goyal2017making}, and GQA~\citep{hudson2019gqa} have centered on traditional VQA queries. Recently, there are works like MM-Vet~\citep{yu2023mm}, POPE~\citep{li2023evaluating}, and MM-Bench~\citep{liu2023mmbench} designed to specifically evaluate multimodal LLMs including hallucination, reasoning, and robustness. The previous benchmarks and evaluations have shown that Multimodal LLMs can suffer from hallucination~\citep{liu2023aligning,liu2023hallusionbench}, catastrophic forgetting~\citep{zhai2023investigating} and lack of robustness~\citep{fu2023mme}. In taking a step back to the fundamentals, our work uncovers that even the most advanced multimodal LLMs, such as GPT-4V~\citep{gpt4v}, Gemini~\citep{Gemini}, Bard~\citep{liu2023improved}, and LLaVA-1.5~\citep{liu2023improved}, are not immune to stumbling over elementary visual questions. We also identified part of the problem as being the incapable visual encoder.

\thinparagraph{Visual Encoders.}
\ours{}-VLM provides a detailed analysis of the visual capabilities of various CLIP variants~\citep{radford2021learning, sun2023eva, xu2023demystifying, zhai2023sigmoid}. These models mostly follow the method proposed in \citet{radford2021learning} that uses contrastive loss to train on large volumes of image-text pairs. They differ in training data~\citep{xu2023demystifying}, training recipes~\citep{sun2023eva}, and objective functions~\citep{zhai2023sigmoid}. Nonetheless, our studies show that all of these CLIP variants struggle with simple visual patterns such as ``orientation'', ``count'', ``presence of specific features'', \etc. Another line of research focuses on vision-only self-supervised learning (SSL). This category includes contrastive SSL~\citep{chen2020simple, grill2020bootstrap, bardes2021vicreg, he2020momentum} and mask-based SSL~\citep{zhou2021ibot, he2022masked, assran2023self}. SLIP~\citep{mu2022slip} explores the synergy between CLIP and contrastive SSL, but focusing primarily on standard classification tasks. In fact, a common practice to evaluate the quality of these vision models is through linear probing or fine-tuning on ImageNet ~\citep{russakovsky2015imagenet, ridnik2021imagenet}. Although current evaluation methods provide a basic level of assessment on representation quality, our findings indicate a growing detachment from the needs of recent use cases. As demonstrated in the MoF experiments in Section \ref{sec: incorporate_dino}, the CLIP vision model and the vision-only SSL models learn complementary features. However, the linear probing accuracy on ImageNet alone provides a limited understanding of feature utility in MLLMs. This observation suggests the need for more diverse evaluations~\citep{vishniakov2024convnet} in visual representation learning, to better align with current and emerging applications.

\thinparagraph{Ambiguities in Embedding Models.}
Our work exploits CLIP-blind pairs within the CLIP vision embedding space to generate examples of failures in CLIP models and subsequently MLLMs. This concept has ties to previous research focused on documenting failure modes in text embedding models~\citep{gonen2019lipstick, may2019measuring, sun2019mitigating}. More recently, \citet{thrush2022winoground}, \citet{yuksekgonul2022and} and \citet{hsieh2023sugarcrepe} study the binding problems CLIP faces in processing text queries, noting that CLIP models treat text input as a bag of words. \citet{tong2023mass} examines the implications for downstream text-guided generative models. \citet{tschannen2023image} suggests image captioners as promising alternatives to CLIP for improving attribute binding. Our work focuses on the visual patterns.

%% file: sections/6_discussion.tex
\section{Discussion}
Circling back to the very first question we ask: is vision good enough for language? Perhaps not yet, as our study shows that vision models might become a bottleneck in multimodal systems.  MLLMs fail in simple questions because their pre-trained CLIP vision encoders overlook crucial visual details in images, and systematically fail to sort important visual patterns. Yet, CLIP-type models remain the most scalable and widely used vision models today. Contrary to the popular belief that data and model scaling is a panacea, our research demonstrates that scaling alone does not rectify the inherent deficiencies in CLIP models.

Our study reveals that popular visual representation learning models -- vision-and-language models and vision-only self-supervised learning models -- excel in different aspects. The distinction in their capabilities go beyond conventional benchmarks such as linear probing or zero-shot accuracy on ImageNet. Although a carefully designed Mixture-of-Features approach could alleviate visual limitations and utilize the strengths of these two learning paradigms, it is necessary to develop new evaluation metrics to facilitate the development of new visual representation learning algorithms. We hope our work can motivate further innovation in vision models.\vspace{0.5em}

\clearpage

%% file: sections/7_appendix.tex
\clearpage
\appendix
\section{Experiment Details} \label{appendix: experiment details}
\paragraph{Hyperparameters.}
In this work, we adopt the same set of hyperparameters as LLaVA~\citep{liu2023visual} and LLaVA-1.5~\citep{liu2023improved}. We use Vicuna-13b-v1.3~\citep{zheng2023judging} in LLaVA experiments and Vicuna-13b-v1.5~\citep{zheng2023judging} in LLaVA-1.5 experiments. We show the training hyperparameters for LLaVA and LLaVA-1.5 experiments in Table~\ref{tab:hyperparameters}.  All experiments are conducted using a maximum of 8 Nvidia A100 GPUs.

\begin{table}[h]
\centering
    \small

\begin{tabular}{@{}lcccc@{}}
\toprule
\multirow{2}{*}{Hyperparameter} & \multicolumn{2}{c}{LLaVA} & \multicolumn{2}{c}{LLaVA-1.5} \\
                                & Stage 1        & Stage 2       & Stage 1         & Stage 2        \\ \midrule
batch size                      & 128            & 128           & 256             & 128            \\
lr                              & 1e-3           & 2e-5          & 2e-3            & 2e-5           \\
lr schedule decay                    & cosine   & cosine   & cosine                & cosine               \\
lr warmup ratio                 & 0.03           & 0.03          & 0.03               & 0.03              \\
weight decay                    & 0              & 0             & 0               & 0              \\
epoch                           & 1              & 3             & 1               & 1              \\
optimizer                       & \multicolumn{4}{c}{AdamW~\citep{loshchilov2017decoupled}}                     \\
DeepSpeed stage                 & 2              & 3             & 2               & 3              \\
\bottomrule
\end{tabular}
\caption{Hyperparameters for MoF training on LLaVA and LLaVA-1.5.}
\label{tab:hyperparameters}
\end{table}

\paragraph{Pretrain Datasets.}
We use the same dataset for both LLaVA and LLaVA-1.5 experiments. For LLaVA experiments, stage 1 uses CC595k~\citep{sharma2018conceptual} and stage 2 uses LLaVA 158k~\citep{liu2023visual} instruction data; For LLaVA-1.5 experiments, stage 1 uses CC595k~\citep{sharma2018conceptual} and stage 2 uses DataMix 665k~\citep{liu2023visual, sharegpt, goyal2017making, hudson2019gqa, marino2019ok, mishra2019ocr, schwenk2022okvqa, sidorov2020textcaps, mao2016generation, kazemzadeh2014referitgame, krishna2017visual} proposed in \citet{liu2023improved}. 

\section{\ours{} Benchmark} 
We provide more details on the \ours{} benchmark. 
\subsection{Details of evaluating SOTA models} \label{appendix: access the model}

We access GPT-4V through ChatGPT in October and November 2023. We also evaluate Gemini-Pro through Vertex AI API in December 2023. We use the official checkpoints for InstructBLIP~\citep{instructblip}. We access mini-GPT4~\citep{zhu2023minigpt},\footnote{To circumvent response hallucination in mini-GPT4 we prefix our questions with ``Please only choose an option to answer the question below without explanation: ''} LLaVA and LLaVA-1.5~\citep{liu2023visual} through their playgrounds. We test Bard~\citep{Bard} using the official website in September and October 2023. Moreover, we test new-Bing \cite{newbing} through new-Bing chat creative mode and GPT-4V \citep{gpt4v} in September 2023.

\subsection{Questions in \ours{} Benchmark} \label{appendix: full benchmark}

\begin{figure*}[t]
  \centering
  \includegraphics[width=0.98\textwidth]{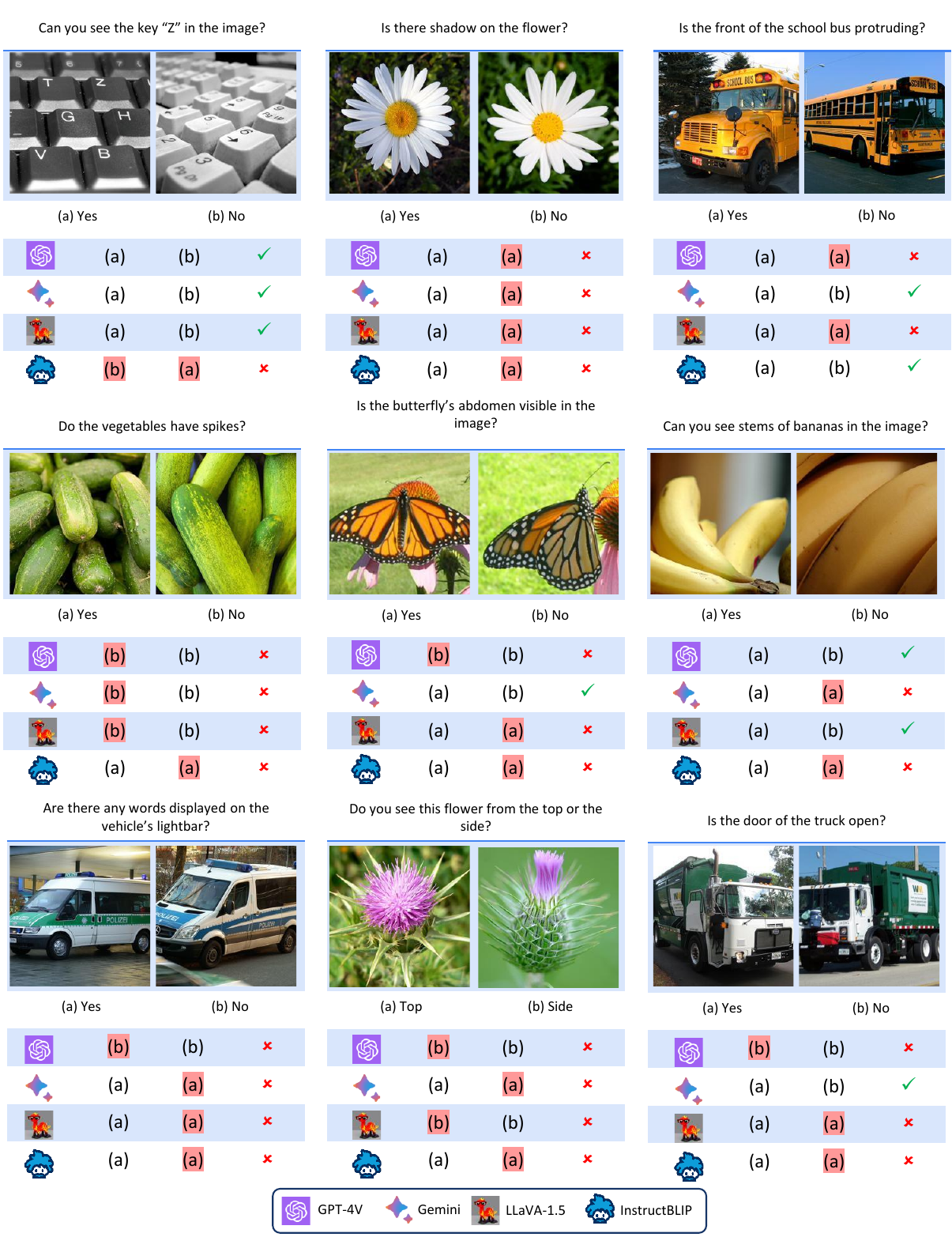}
  \caption{\textbf{More examples of questions in the \ours{} benchmark (Part I).}}
  \label{fig:more example_questions I}
  \vspace{-0.5cm}
\end{figure*}

\begin{figure*}[t]
  \centering
  \includegraphics[width=0.98\textwidth]{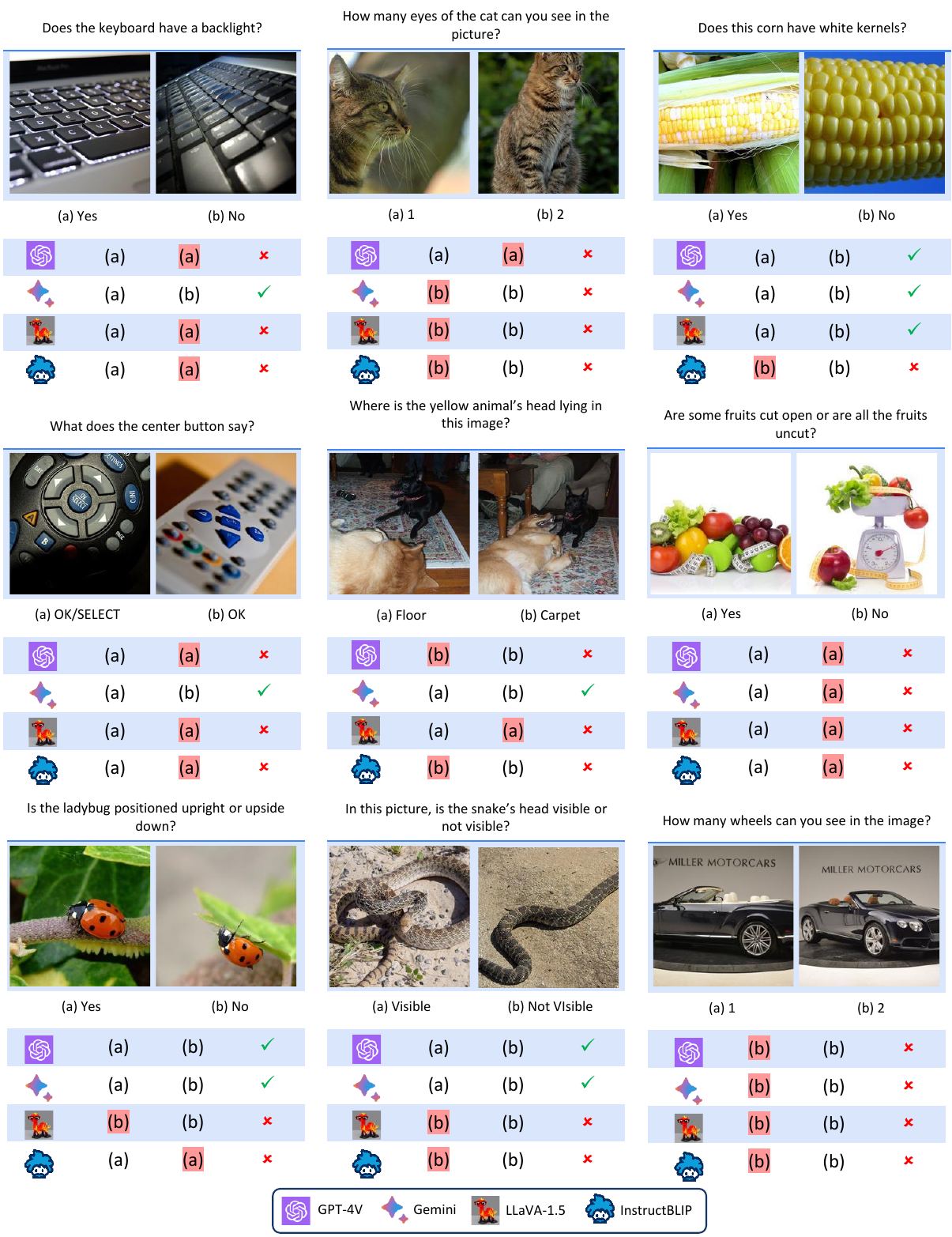}
  \caption{\textbf{More examples of questions in the \ours{} benchmark (Part II).}}
  \label{fig:more example_questions II}
  \vspace{-0.5cm}
\end{figure*}

\begin{figure*}[t]
  \centering
  \includegraphics[width=0.98\textwidth]{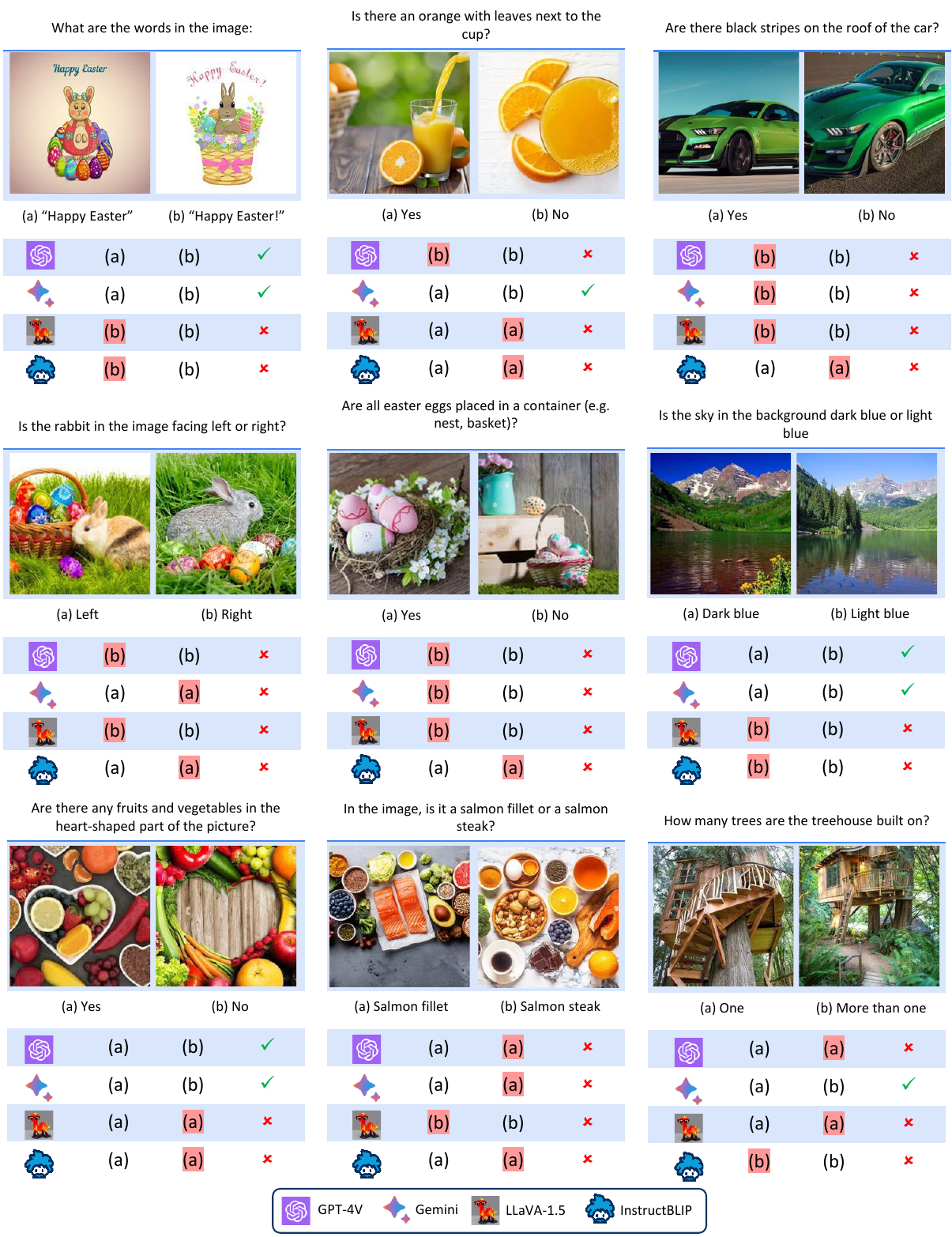}
  \caption{\textbf{More examples of questions in the \ours{} benchmark (Part III).}}
  \label{fig:more example_questions III}
  \vspace{-0.5cm}
\end{figure*}

We present more examples in \ours{} at the end in Figures~\ref{fig:more example_questions I},~\ref{fig:more example_questions II},~\ref{fig:more example_questions III}.

\subsection{Ablation Studies} \label{appendix: ablation study on questions}
To further verify that MLLMs make mistakes in \ours{} due to their incapable visual grounding instead of hallucination in the language model~\citep{hu2023prompt}. We conduct additional ablation experiments on the format and notations of VQA questions and options in \ours{}. We choose GPT-4V to do these experiments, as it is currently the best model. 
\paragraph{Swapping options}
The first experiment swaps the two options in the \ours{} benchmark. For example, we change the question from ``Are the butterfly's wings closer to being open or closed?	(a) Open (b) Closed'' to  ``Are the butterfly's wings closer to being open or closed?	(a) Closed (b) Open''.

Empirically, we find that GPT-4V obtains a 40.3\% accuracy on the option swapping in our study, as opposed to the original 38.7\%. We observe that a few questions are answered differently, while the majority remain the same. This further suggests that the visual incapabilities are in the vision encoder rather than in alignment or the LLMs.

\paragraph{Changing notations in the options}
We conducted an ablation study to assess the impact of altering notations. For example, we changed ``(a) Closed (b) Open'' to ``(1) Closed (2) Open''. The results are comparable to the original findings, achieving a performance of 37.3\%, closely matching the original 38.7\%. The study further suggests that the core challenge in MLLMs is their inherent visual incapability, rather than hallucinations in the language model.

\subsection{Human Study Details} \label{appendix: human study}
\begin{figure*}[t]
    \centering
  \includegraphics[width=0.99\textwidth]{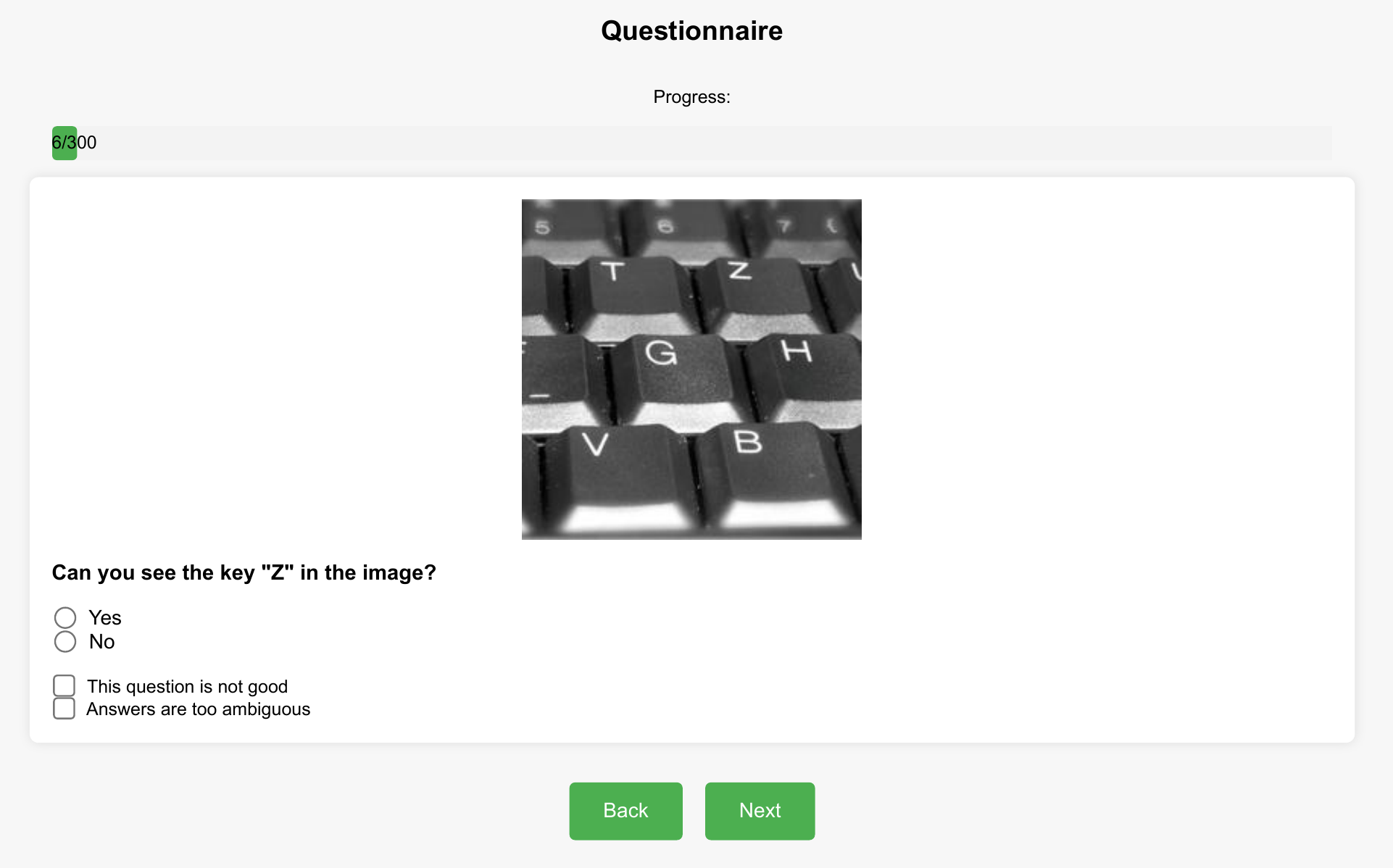}
  \caption{\textbf{Example of user study interface.} The questions in the user study are randomly shuffled to avoid any potential bias. Users choose answers for the VQA questions as well as potential concerns for the VQA question. }%\pt{Image text are too repetitive or show from wrong to right}}
  \label{fig: User Interface}
  \vspace{-0.5cm}
\end{figure*}

In this study, we ask four participants to volunteer in our study. An example user interface for labeling is shown in Figure~\ref{fig: User Interface}. We collect their responses and calculate the average score as the human-level performance.

\section{CLIP-MLLM Failure Correlation}  \label{appendix: correlation}
\paragraph{Correlation between CLIP and MLLM models.}
We compute the Pearson Correlation between the CLIP model and MLLMs and show results in Table~\ref{tab:correlation}. Notably, both open-source models -- LLaVA and InstructBLIP -- exhibit remarkably high Pearson Correlation, exceeding 0.7. This finding indicates a strong correlation between the errors made by the CLIP model and those made by MLLMs. Bard also displays a very high correlation. This suggests that some of the most advanced closed-source models are also affected by the visual limitations in the CLIP models.

\begin{table}
\setlength\tabcolsep{3pt}
\centering
\small
\begin{tabular}{cccccc}
\toprule
&LLaVA-1.5 & InstructBLIP & Bard & Gemini & GPT-4 \\
\midrule
Correlation & \cellcolor{lightgray} 0.87   & \cellcolor{lightgray} 0.71            & 0.79       & 0.72    & 0.31                   \\
\bottomrule
\end{tabular}
\caption{Pearson Correlation between the CLIP model and MLLMs. Open-source models that explicitly use CLIP-based models are highlighted in \colorbox{lightgray}{gray.}}
\label{tab:correlation}
\end{table}

\begin{figure}[t]
    \centering
  \includegraphics[width=0.49\textwidth]{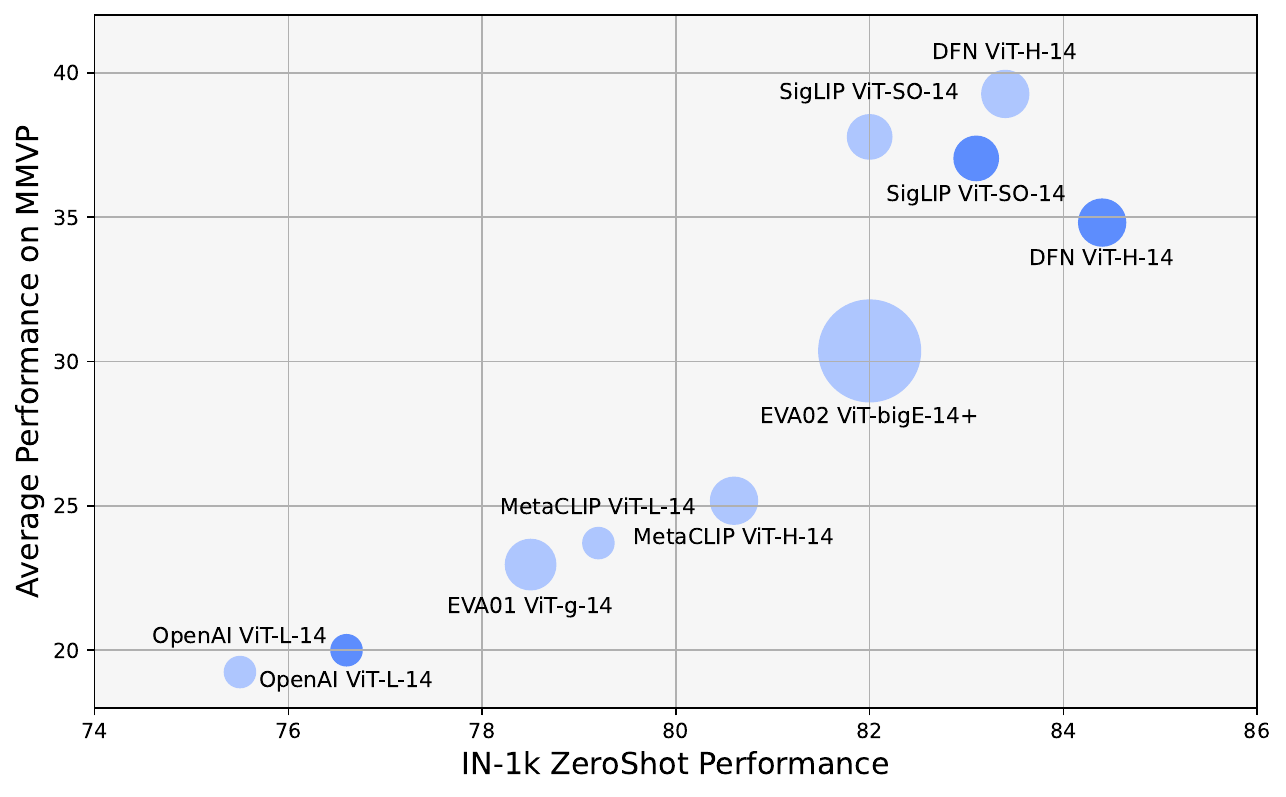}
  \caption{\textbf{Correlation between ImageNet-1k Zero-shot and \ours{}-VLM average.} The area of each bubble corresponds to the model's number of parameters. A higher ImageNet-1k zero-shot performance does not necessarily imply superior performance in \ours{}-VLM.}%\pt{Image text are too repetitive or show from wrong to right}}
  \label{fig: in1kmmvp correlation}
  \vspace{-0.5cm}
\end{figure}

\paragraph{Correlation between ImageNet-1k and \ours{} performance.}\
We plot the ImageNet-1k Zero-shot accuracy against \ours{}-VLM average performance in Figure~\ref{fig: in1kmmvp correlation}. For models with ImageNet-1k Zero-shot accuracy below 80, a higher Zero-shot accuracy tends to indicate improved \ours{} performance. However, in models with superior ImageNet-1k Zero-shot performance, this trend does not necessarily hold for \ours{}-VLM accuracy. This distinction accentuates the value of \ours{}-VLM as an evaluation metric, which probes into visual patterns such as orientation – aspects that are pivotal for downstream tasks and go beyond what is captured by ImageNet accuracy alone. 

\section{Visual Patterns for CLIP} \label{appendix: visual patterns}
Here, we provide the full description of visual patterns that pose challenges to all CLIP-based models. 

\begin{itemize}
    \item \textbf{\faCompass \text{ }Orientation and Direction}: Questions about the direction something is facing or moving, such as the direction the dog or duck is facing, or the orientation of the school bus.
    \item \textbf{\faSearch \text{ }Presence of Specific Features}: Questions that focus on the existence or non-existence of certain elements or features in the image.
    \item \textbf{\faSync\text{ } State and Condition}: Questions that pertain to the state or condition of an object, such as whether a flag is blowing in the wind or if the ground is wet.
    \item \textbf{\faSortNumericUp \text{ }Quantity and Count}: Questions about the number of objects or features present in the image.
    \item \textbf{\faMapPin\text{ } Positional and Relational Context}: This aspect refers to the model's ability to understand the position and relationship of objects or elements within an image in relation to each other and their surroundings. 
    \item \textbf{\faPalette\text{ } Color and Appearance}: Questions regarding the color of certain objects or elements.
    \item \textbf{\faCogs\text{ } Structural and Physical Characteristics}: This category involves the model's ability to identify and analyze the physical attributes and structural features of objects in an image. 
    \item \textbf{\faFont\text{ } Text}: Questions related to text or symbols present in the image.
    \item \textbf{\faCamera \text{ }Viewpoint and Perspective}: Questions concerning the perspective from which the photo was taken.

\end{itemize}

\section{More Benchmark Results}

\subsection{Different vision-only backbones} \label{appendix: mae&moco}
Here, we conduct extra experiments to study MoF involving MAE~\citep{he2022masked} or MoCoV3~\citep{he2020momentum} instead of DINOv2; See Table~\ref{tab:different visual encoder}. In Table~\ref{tab:different visual encoder}, we observe that with MAE/MoCov3, there is a consistent improvement in visual grounding ability, as shown in the \ours{} and POPE benchmarks.
\begin{table}[h!]
    \centering
    %\vspace{-0.3cm}
    \small
    \setlength\tabcolsep{2.5pt} % Default value: 6pt
    \begin{tabular}{l c c c l l l}  
        %\toprule
         method & SSL Model & res & \#tokens & MMVP  & POPE
        \\
        \toprule
        LLaVA$^{1.5}$  & None & 336$^2$ & 576 & 24.7 & 85.9\\
        LLaVA$^{1.5}$ + I-MoF & MoCov3 & 224$^2$ & 512 & 26.7{\textbf{\tiny{\plusvalue{2.0}}}} & 86.1  \\
        LLaVA$^{1.5}$ + I-MoF & MAE & 224$^2$ & 512 & 27.3{\textbf{\tiny{\plusvalue{2.6}}}} & 86.1
        \\
        LLaVA$^{1.5}$ + I-MoF & DINOv2 & 224$^2$ & 512 & 28.0{\textbf{\tiny{\plusvalue{3.3}}}} & 86.3
    \end{tabular}
    %\vspace{-0.4cm}
    \caption{\footnotesize Results of Interleaved MoF with different vision-only SSL model}
    %\vspace{-0.4cm}
    \label{tab:different visual encoder}
\end{table}

\subsection{Scaling up to larger resolution}  \label{appendix: more benchmarks}

\begin{table*}[h]
    \centering
    \begin{tabular}{l c c l l l l l l l l}  
         method  & res & \#tokens & MMVP & LLV$^{\text{B}}$ & LLV$^\text{W}$  & MMB & VQA$^\text{T}$  & POPE & VQA$^\text{V2}$  & MM-V
        \\
        \toprule

        LLaVA$^{1.5}$  & 336$^2$ &\cellcolor{Green} 576 & 24.7 &  \textbf{84.7} & 70.7 & \textbf{67.7} &  \textbf{61.3} & 85.9 & \textbf{80.0} & \textbf{35.4}\\
         LLaVA$^{1.5}$ + I-MoF  & 224$^2$ & \cellcolor{MidGreen}512 &  28.0 & 82.7 & \textbf{73.3}  & 61.6 & 55.3& 86.3& 77.3 & 33.5 \\
         LLaVA$^{1.5}$ + I-MoF  & 336$^2$ & \cellcolor{DarkGreen}1152 & \textbf{31.3} & 81.8 & \textbf{73.3}  & 65.4 & 58.7 & \textbf{86.7} & 79.3 & 34.6\\
    \end{tabular}
    \caption{\textbf{Comparison with LLaVA-1.5 on 6 more benchmarks}. Interleaved-MoF LLaVA-1.5 obtains performance on par with the original method while showing improvements on benchmarks evaluating visual grounding.
    Benchmark names are abbreviated due to space limits. LLV$^\text{B}$: LLaVA Benchmark~\citep{liu2023visual}; LLV$^\text{W}$: LLaVA-In-the-Wild~\citep{liu2023improved}; MMB: MMBench~\citep{liu2023mmbench}; VQA$^\text{T}$: TextVQA\citep{singh2019towards}; POPE: POPE~\citep{li2023evaluating}; VQA$^\text{V2}$: VQA-v2~\citep{goyal2017making}; MM-V: MM-Vet~\citep{yu2023mm}.} 
    \label{tab:More Benchmark}
\end{table*}

We conduct additional experiments on Interleaved-MoF that further scale up the resolution to 336 and evaluate on more benchmarks. The summarized results in Table~\ref{tab:More Benchmark} reveal that Interleaved-MoF achieves comparable performance on most benchmarks while demonstrating improvements in benchmarks focused on visual grounding. We also observe that \ours{} are more sensitive to the model's visual capabilities, underscoring the significance of our benchmark in assessing visual proficiency.

